\documentclass[sn-mathphys]{sn-jnl}

\jyear{2021}%

\theoremstyle{thmstyleone}%
\theoremstyle{thmstyletwo}%

\theoremstyle{thmstylethree}%

\usepackage{mathrsfs}
\usepackage{amsmath}

\begin{document}

\title[A Review on Machine Unlearning]{A Review on Machine Unlearning}


\author*[1]{\fnm{Haibo} \sur{ZHANG}}\email{zhang.haibo892@s.kyushu-u.ac.jp}

\author[2]{\fnm{Toru} \sur{NAKAMURA}}\email{tr-nakamura@kddi-research.jp}

\author[2]{\fnm{Takamasa} \sur{ISOHARA}}\email{ta-isohara@kddi-research.jp}

\author[3]{\fnm{Kouichi} \sur{SAKURAI}}\email{sakurai@inf.kyushu-u.ac.jp}

\affil*[1]{\orgdiv{Department of Information Science and Technology}, \orgname{Graduate School of Information Science and Electrical Engineering}, \orgaddress{\street{Kyushu University}, \city{Japan}, \postcode{819-0395}}}

\affil[2]{\orgdiv{KDDI Research Inc.}, \city{Japan}, \postcode{356-8502}, }


\affil[3]{\orgdiv{Department of Information Science and Technology}, \orgname{Faculty of Information Science and Electrical Engineering}, \orgaddress{\street{Kyushu University}, \city{Japan}, \postcode{819-0395}}}


\abstract{Recently, an increasing number of laws have governed the useability of users’ privacy. For example, Article 17 of the General Data Protection Regulation (GDPR), \textit{the right to be forgotten}, requires machine learning applications to remove a portion of data from a dataset and retrain it if the user makes such a request. Furthermore, from the security perspective, training data for machine learning models, i.e., data that may contain user privacy, should be effectively protected, including appropriate erasure. Therefore, researchers propose various privacy-preserving methods to deal with such issues as machine unlearning. This paper provides an in-depth review of the security and privacy concerns in machine learning models. First, we present how machine learning can use users' private data in daily life and the role that the GDPR plays in this problem. Then, we introduce the concept of machine unlearning by describing the security threats in machine learning models and how to protect users' privacy from being violated using machine learning platforms. As the core content of the paper, we introduce and analyze current machine unlearning approaches and several representative research results and discuss them in the context of the data lineage. Furthermore, we also discuss the future research challenges in this field.}




\maketitle

\section{Introduction and Background}

Privacy protection has been a concern for researchers for a long time. In today's big data environment, users interact with data on various web platforms, such as sending and receiving emails, and browsing news, almost every day. For users, once they have provided their information in an application, it is difficult to remove it from the root. When machine learning is widely used today, most advanced features are obtained based on understanding and training data. As a result, users' privacy has been spread in every corner of the application, makings it more accessible for attackers to steal users' private data.

From the security perspective, if an attacker compromises the machine learning model by injecting some pollution data into its dataset, it is also necessary to remove such data from the dataset and retrain it \cite{baracaldo2017mitigating}. For example, an attacker can open a backdoor in a machine learning model by injecting malicious data into the dataset used for training \cite{liu2022backdoor}. As a result, the attacker can steal all the private data in the model, shown in Figure 1.

\begin{figure*}[!htpb]
\centerline{\includegraphics[scale=0.55]{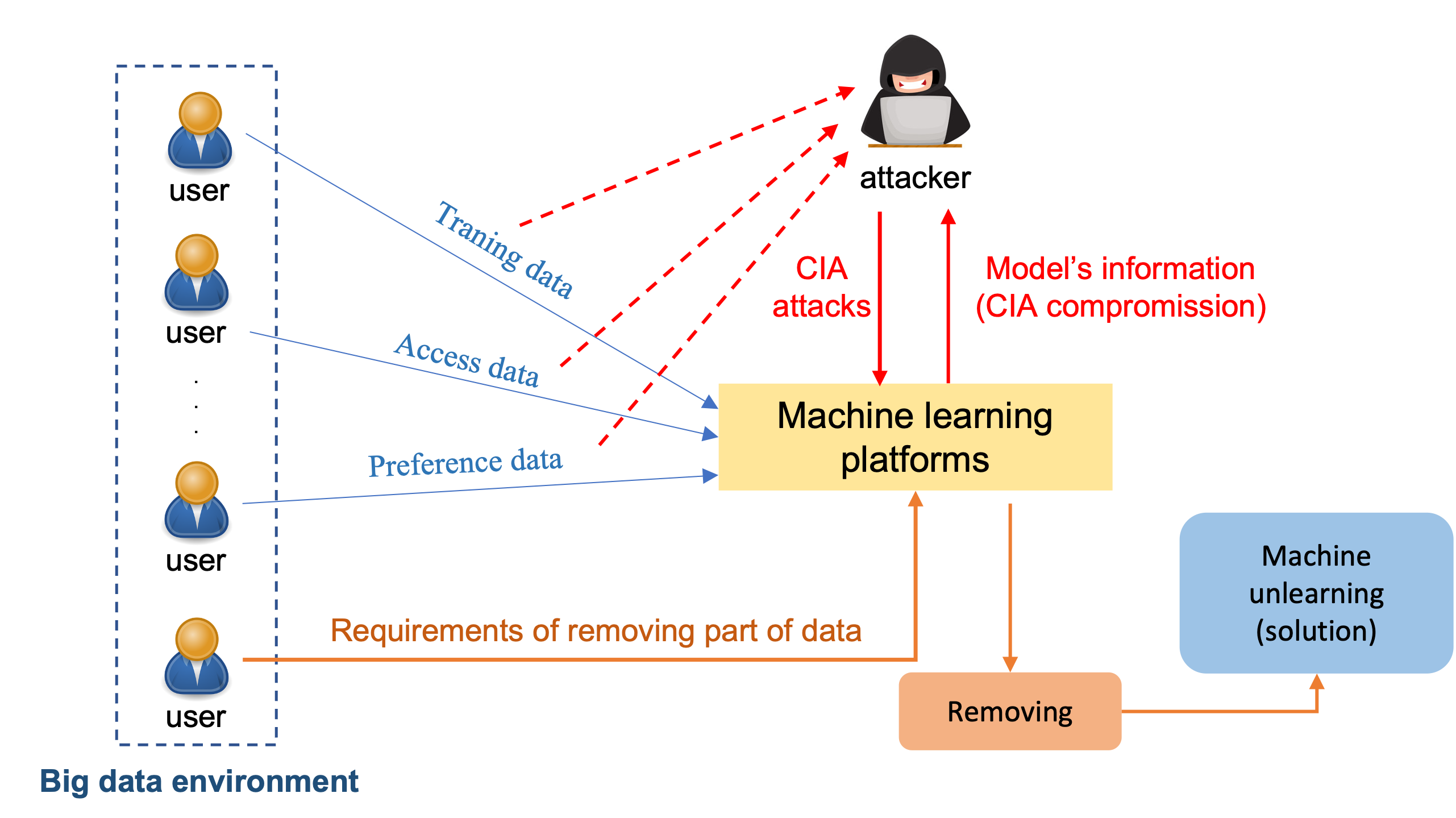}}
\caption{The necessity of machine unlearning. The red arrow indicates that the attacker can access the training data or parameters of the machine learning model through malicious data injection or information stealing to obtain user privacy or even reconstruct the machine learning model. In this case, according to the orange arrow, the data owner will request to delete specific sensitive data, and the model owner needs to apply machine unlearning methods to remove the requested data.}
\label{fig1}
\end{figure*}

For solving the above problems, it is necessary to retrain the machine learning model. However, the existing retraining methods cause a large amount of computational power and time consumption. Therefore, researchers propose machine unlearning as a more efficient research method \cite{bourtoule2021machine}. 

The word “unlearning” means that the machine learning model is re-trained to generate a new predictive model with a portion of the data forgotten. There are two ways to perform unlearning on machine learning models. One is to retrain the new dataset from scratch after data removal (i.e., exact unlearning mentioned in Section 4.4. The other is to modify the machine learning model and dataset to achieve an approximate unlearning effect (i.e., approximate unlearning mentioned in Section 4.5). The ultimate goal of either unlearning approach is to improve the accuracy of unlearning methods while being as efficient as possible.

This paper provides an in-depth analysis of machine learning models' security and privacy concerns, which also refers to the privacy-preserving machine learning \cite{al2019privacy}. This paper aims to provide a comprehensive analysis and summary of current machine unlearning techniques and future research potential.

First, we present how machine learning can use users' private data in daily life and the role that the GDPR plays in this problem. Then, we introduce the concept of machine unlearning by describing the security threats in machine learning models and how to protect users' privacy from being violated using machine learning platforms. In the next section, we introduce and analyze current machine unlearning approaches and several representative research results and discuss them in the context of the data lineage. Furthermore, we also discuss the future research challenges in this field.

\subsection{How Machine Learning can use users' data?}
Since the idea of simulating human intelligence was first proposed in the 1960s, artificial intelligence (AI) received widespread attention in both academical and industry fields. As the primary component of AI, machine learning also gained unprecedented development in recent years. Moreover, its application has spread to various fields of AI. For example, we can use machine learning to classify and locate objects in the field of computer vision, and we can also use deep neural networks to design and implement a high-accuracy face recognition system. In addition, we can also use machine learning in natural language processing to design and implement an intelligent question and answer system.

In the modern Big Data environment, Internet users interact with various applications almost every day. Enterprises and developers use data mining, big data analytics, and machine learning techniques to extract useful information from the vast database. This data contains more or less sensitive information about users, such as their identity and passwords. Hence, machine learning plays an important role.

Machine learning is a branch of artificial intelligence that automatically enables computers to learn from experience through human intervention. The whole concept of machine learning starts around determining the answer to an obstacle without human interference, which begins with understanding data from examples or direct experience, analyzing data patterns and making better decisions based on inferences. It is best used for problem-solving when large amounts of data and variables exist without using existing algorithms. For example, Google tends to optimize search results and pop-up ads for products similar to users' tastes or websites they have visited before. It studies the user's behavior and displays the results accordingly.

Machine learning is an integral part of big data analytics. Big data analytics includes big data, data learning, statistical information, etc. Machine learning uses programming and computational algorithms to conclude, while big data analytics uses numbers and statistics to draw results.

\subsection{The General Data Protection Regulation}
Recently, an increasing number of laws have governed the usability of users’ privacy. For example, Article 17 of the General Data Protection Regulation (GDPR), \textit{the right to be forgotten}, requires entities to remove a portion of data from a dataset if the user makes such a request \cite{scheltertowards,graves2020amnesiac}. Furthermore, it maintains the user's right to use their privacy from a privacy protection perspective \cite{al2019privacy}.

The GDPR is a new EU privacy and data protection regulation. It requires more granular privacy protections in company systems, more detailed data protection agreements, and more user-friendly and detailed disclosures about company privacy and data protection practices.

The GDPR has direct legal implications for all EU member states, i.e., it is binding without having to be transposed into the national laws of EU member states. This will enhance the consistency and harmonization of the implementation of EU law.

From its initial draft in 2012 to becoming official EU law in 2016, \textit{the right to be forgotten} was initially intended to bind Internet search engines, such as Google and Yahoo, in their use of users' privacy. Under the Article 17, if a user requests the deletion of any private data, the search engine shall immediately execute and is not allowed to refuse. However, implementing this law also raises considerations about the current hot topic, machine learning technology, and overusing users' private data. For example, how should machine learning platforms respond if data holders request to delete specific data used for training purposes?

In the context of machine learning, \textit{the right to be forgotten} requires the machine learning applications to be able to readily accommodate requests from data owners who wish to delete any data \cite{chen2021machine}. This process is called machine unlearning. The machine learning application needs to remove the requested data from the training dataset and retrain the machine learning model from scratch.

The appearance of the Article 17 has primarily limited the undesirable phenomenon of misused and unprotected user privacy in the current fast-growing Internet and big data environment. However, privacy protection should be carried out from both the perspective of the data controller and the data holder. \textit{The right to be forgotten} can be regulated from the perspective of data controllers, but it does not work from the perspective of data holders. That is, how data holders become aware of the violation of their privacy and when they request the deletion of their private data. These cannot be regulated by the regulation and require data holders to raise their awareness of privacy protection under the guidance of social engineering.

\section{Security Concerns}
\subsection{Machine Learning is Still Weak}
In the era of big data and artificial intelligence, people can access information more quickly and efficiently. However, while gaining convenience, our behavior is being recorded, learned and used all the time. If we ignore privacy protection in the application, it will be challenging to prevent personal information from being used for illegal purposes.

Due to the vulnerability problem of machine learning models themselves, attackers can attack machine learning models by sending many malicious requests, exposing machine learning services to various potential security risks \cite{gao2022deletion}.

\begin{itemize}
    \item Data privacy leakage risk: Attackers can exploit the model vulnerability to obtain data information for training models by invoking machine learning services.
    \item Model theft risk: model parameter information in machine learning services due to the model's vulnerability issues, making it risky for attackers to speculate and restore model parameter information by frequently invoking the service.
    \item Data Poisoning: The attacker can mix specific malicious data in the request process, which can affect the model training and subsequent model inference through the feedback of the service process to achieve the effect of interfering with the model \cite{marchant2021hard,baracaldo2018detecting}.
    \item Evasion: Attackers can make machine learning services make wrong judgments by adding a small amount of noise and perturbation to typical requests.
\end{itemize}

Usually, when designing machine learning systems, developers consider specific threat models to ensure that the designed system is secure and trustworthy. So far, most of the existing machine learning models have been designed and implemented for a fragile threat model without much consideration of the attackers \cite{chundawat2022zero}. Although these models can perform very well in the face of natural inputs, in a realistic setting, these machine learning models encounter many malicious users and even attackers.

Toreini et al. \cite{toreini2020relationship} provides a systematic approach to relate considerations about trust from the social sciences to trustworthiness technologies proposed for AI-based services and products. For example, attackers have different degrees of ability to maliciously modify the inputs and outputs during the model's training and prediction phases. Even they can access the internal structure of the model by some means and steal the parameters, thus destroying the confidentiality, integrity and usability of the models.

\begin{figure*}[!htpb]
\centerline{\includegraphics[scale=0.55]{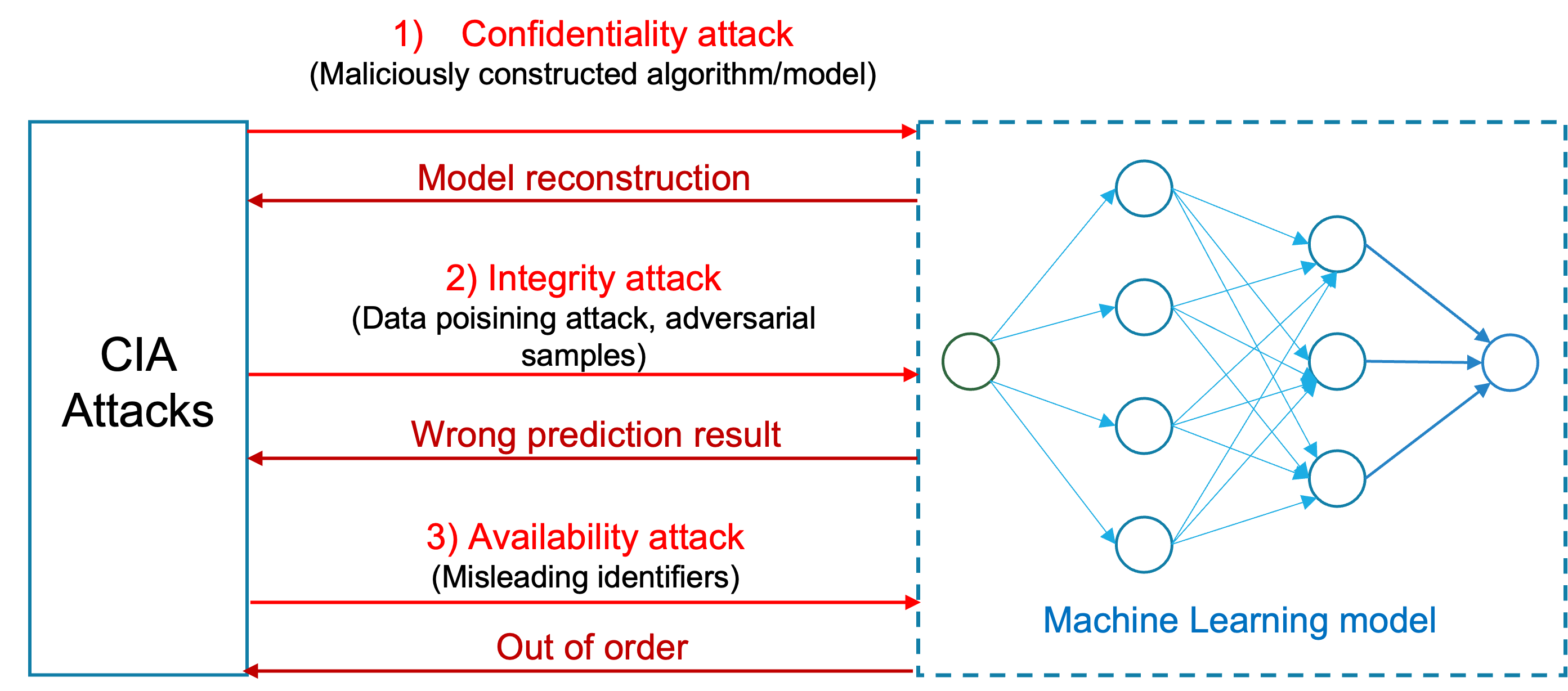}}
\caption{CIA triad in machine learning.}
\label{fig1}
\end{figure*}

\subsection{CIA Triad in Machine Learning}
The CIA triad is a common assessment model that forms the basis for developing security systems and policies. The CIA refers to confidentiality, integrity and availability. The CIA triad identifies system vulnerabilities and methods to address problems and create effective solutions.

Attacks against machine learning models can impact the Confidentiality, Integrity, and Availability \cite{surma2020hacking}. Figure 2 describes how the CIA triad can be applied to the machine learning model.

\begin{itemize}
    \item \textbf{Confidentiality attacks} means that machine learning systems must ensure that unauthorized users do not have access to the information. 
    While most machine learning platforms are professional and secure, the algorithms provided by machine learning model providers are not necessarily reliable \cite{tramer2016stealing}. When data holders use MLaaS (Machine Learning as a Service) to train their predictive models, they may select a malicious model carefully constructed by an attacker. In such models, the attacker encodes the data holder's private data into the parameters of the model and finally steals the user's private data by decoding the parameters of the model \cite{song2017machine}.
    

    \item Machine learning models are most vulnerable to \textbf{integrity attacks}, occurring both in the learning and prediction phases. If the attacker disrupts the model's integrity, then the model's prediction results will deviate from expectations.
    The attacker can modify the existing training set or add additional malicious data to compromise the integrity of the model in order to reduce the accuracy in the prediction phase \cite{shen2016auror}.
    When the model is trained and used for prediction, the attacker only needs to add a small perturbation to the sample to be predicted, which is unrecognizable to the human eye but sufficient to make the model classification wrong.
    
    \item The \textbf{availability} of machine learning models can also be a target of attack. For example, in a driverless scenario, if an attacker places something complicated to identify on the side of the road where a vehicle would pass, it could potentially force a self-driving car to go into safety protection mode and then stop at the side of the road.
\end{itemize}

\section{Privacy-preserving}
In recent years, more and more people have started to pay attention to data privacy and pay more attention to privacy terms when choosing to use client software (apps). Some studies have shown that the protection of privacy can increase the usage rate of users \cite{alsdurf2020covi}.

As research evolves, machine learning models become more powerful and require more training data. For example, some training models in the industry need to use hundreds of gigabytes of data to train billions of parameters. Unfortunately, in many professional fields such as healthcare and financial fraud prevention, data is divided into silos due to privacy or interests, making machine learning face the problem of insufficient valid data. Therefore, information flow and machine learning cannot be achieved without providing guarantees for data privacy.

For privacy-preserving approaches in machine learning, they can be divided into confidential computing, model privacy, and distributed learning \cite{al2019privacy}. 

\begin{itemize}
    \item \textbf{Confidential computing} means that the transmission of data and the computation process is confidential. Current approaches to achieving confidential computing include Multi-party Secure Computation, Homomorphic Encryption and Trusted Executive Environment. Confidential computing can be done to protect data privacy during the training process. So can the trained model cause the leakage of private training data? The answer is yes because machine learning models are overfitted to some extent. The models themselves remember part of the training data, leading to private training data leakage by the published models.
    \item For \textbf{model privacy}, this includes \textbf{differential private machine learning} and \textbf{machine unlearning} algorithms. A common practice to achieve differential privacy is to add noise. Adding noise entails performance loss of the model, and differential privacy machine learning studies how to add noise more economically and how to add the least amount of noise to achieve the best performance for a given privacy loss requirement. Another hot topic of model privacy research is machine unlearning. If implementing differential privacy is viewed as actively designing algorithms to make the output model satisfy the privacy requirements, then machine forgetting is a passive solution to model privacy. It aims to implement the user's "the right to be forgotten" in machine learning models.
    \item The vision of \textbf{federated learning} is to perform multi-party federated machine learning without sharing data, which is essentially a distributed machine learning framework with restricted data access. Compared to classical distributed machine learning, the first layer of constraints in federated learning is data isolation - data is not shared across endpoints, is not balanced, and interactive communication is kept to a minimum.
\end{itemize}

Research on privacy-preserving machine learning has never stopped. Among the many approaches, machine unlearning is emerging and closely related to machine learning algorithms themselves. Current studies on machine unlearning cover various research approaches, such as model privacy, differential privacy, and federation learning. It also demonstrates the importance of machine unlearning in studying privacy-preserving machine learning. Therefore, this paper focuses on machine unlearning as a privacy-preserving approach.

\section{Machine Unlearning}
In this section, we explain the definitions of machine learning and machine unlearning and introduce the two primary approaches of machine unlearning, i.e., \textbf{exact unlearning} and \textbf{approximate unlearning}.

\subsection{Defining Machine Learning}
Machine learning is a technique that makes judgments by predicting possible outcomes. The programmer designs an initial model, trains it on a specific data set, and continuously optimizes the parameters in the model based on the prediction results obtained, which eventually leads to a mature model. Figure 3 shows the general process of a machine learning system.

\begin{figure}[h]
\centerline{\includegraphics[scale=0.5]{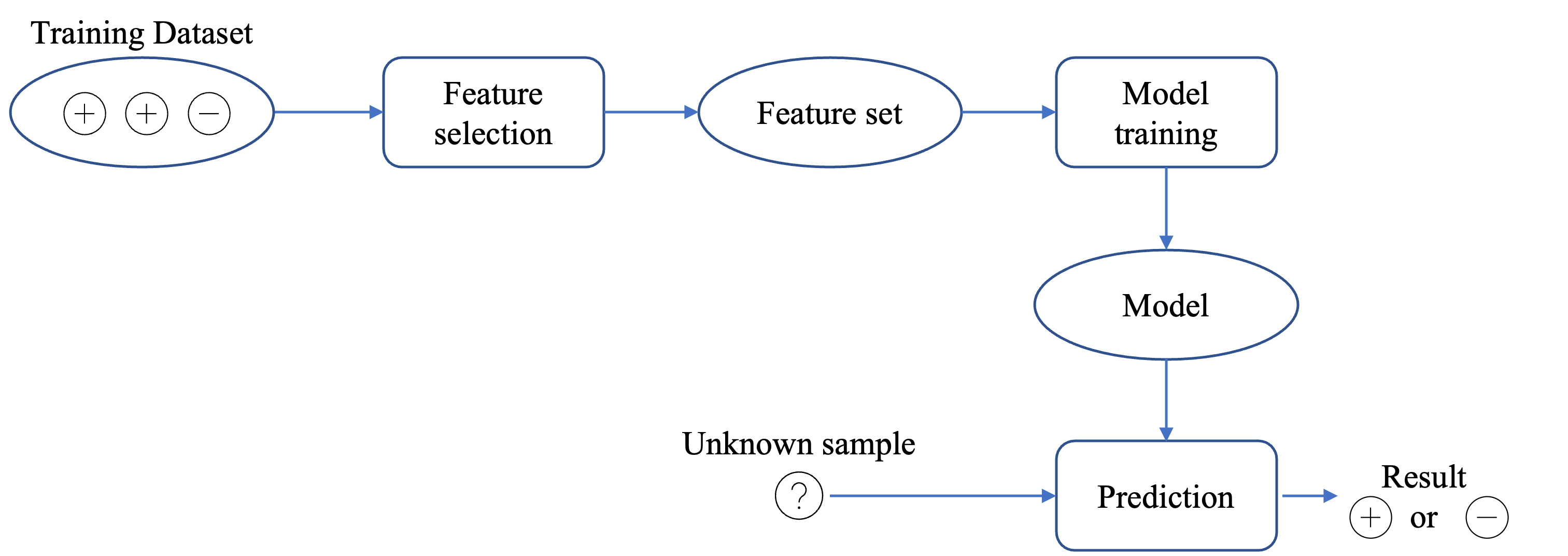}}
\caption{The general machine learning system is consists of three stages, i.e. feature selection, model training and prediction.}
\label{fig3}
\end{figure}

The task to be learned in machine learning can be defined in a space $\mathcal{Z}$ of the form $\mathcal{X}$ $\times$ $\mathcal{Y}$, where $\mathcal{X}$ is named the sample space and $\mathcal{Y}$ is named the output space \cite{bourtoule2021machine}. Taking supervised machine learning as an example, in the image classification problem, for a given data set $\mathcal{D}$ of the input-output pairs (${x}$, ${y}$) $\in$ $\mathcal{X}$ $\times$ $\mathcal{Y}$, the learning aims to find a model function satisfying $\mathcal{F}$: $\mathcal{X}$ $\mapsto$ $\mathcal{Y}$ in a continuous optimization process.

\subsection{Machine Retraining vs. Machine Unlearning}
The most intuitive approach to machine unlearning is to retrain the model on the training data set after deleting the specified data. However, this approach is computationally expensive, so the primary goal of machine unlearning is to reduce the computational cost. One approach is to post-process the trained model so that the results of the machine unlearning algorithm are statistically indistinguishable from the retrained model \cite{ginart2019making}. Another approach is to design new training methods to reduce the cost of retraining. For instance, dividing the data into different blocks, training a separate sub-model for each block, and aggregating the results of the sub-models so that only one sub-model needs to be retrained to remove a data point \cite{bourtoule2021machine,fredrikson2015model}.

\begin{figure*}[!htpb]
\centerline{\includegraphics[scale=0.48]{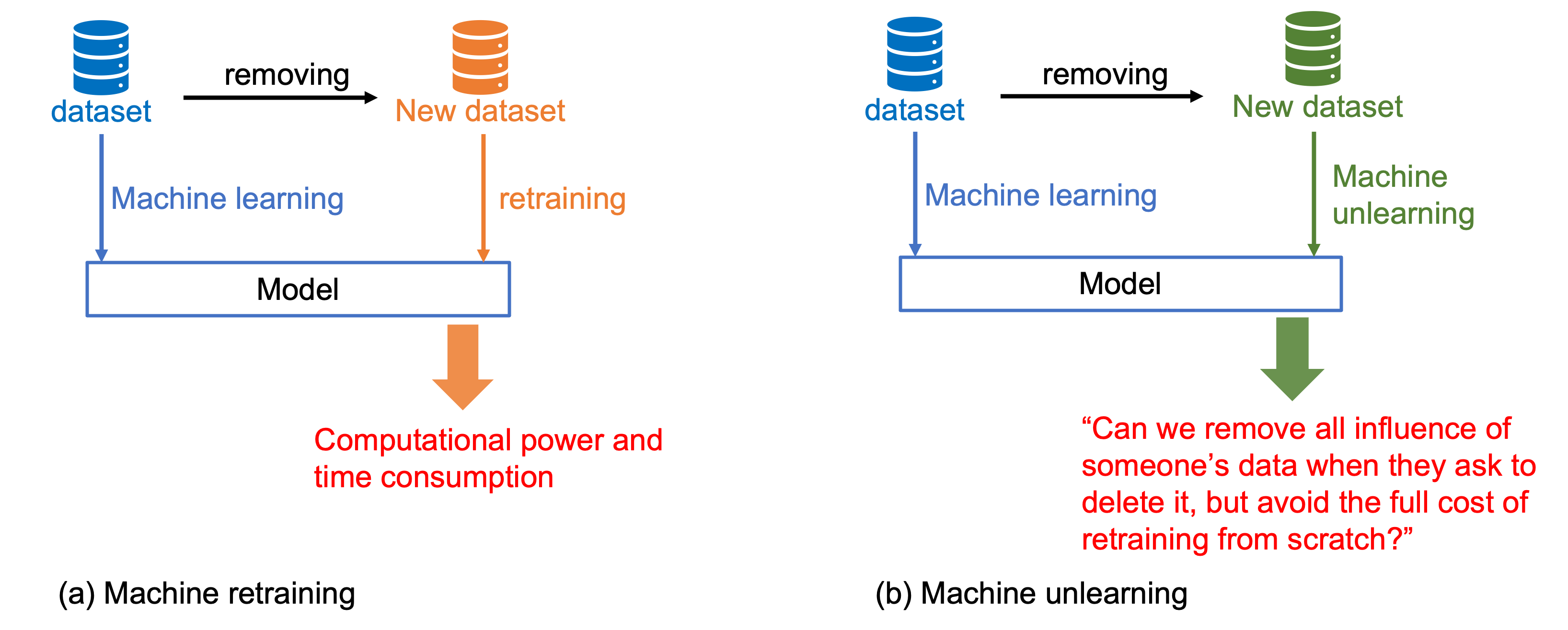}}
\caption{Machine Retraining vs. Machine Unlearning}
\label{fig1}
\end{figure*}

Figure 4 explains the difference between machine retraining and machine unlearning methods. As opposed to removing data from the data set and retraining the entire model, the purpose of machine unlearning is to minimize the cost of time and computational power associated with retraining.

\subsection{Defining Machine Unlearning}
The purpose of machine unlearning is that when the user requests to delete a part of the data, the model that has been learned needs to be retrained in order to generate a model distribution as if that part of the data had not been learned from the beginning.

\begin{figure}[h]
\centerline{\includegraphics[scale=0.5]{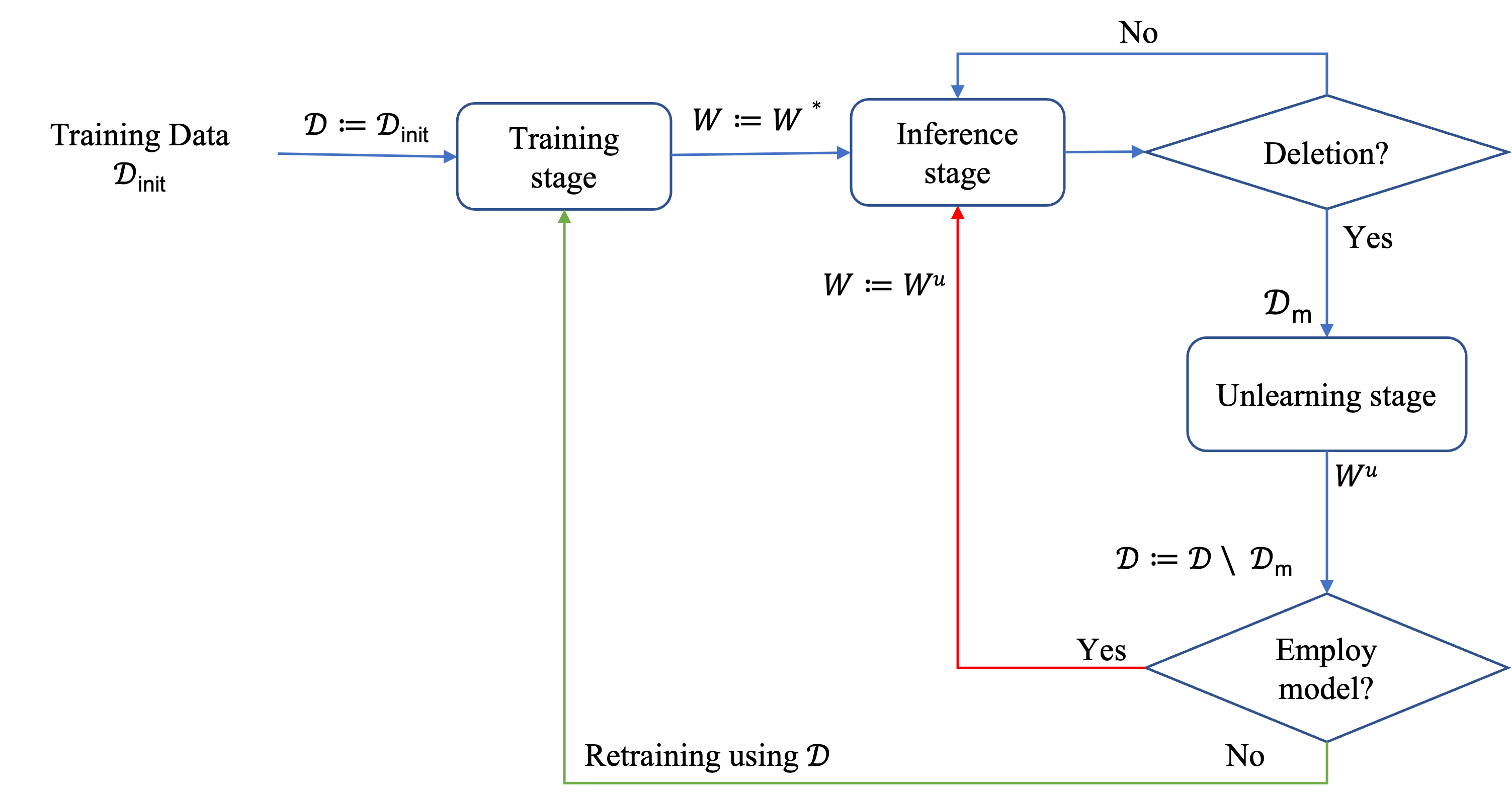}}
\caption{A typical machine learning pipeline consists of three primary stages, i.e., training, inference, and unlearning. First, the initial model $\textit{W}^*$ is trained on the initial dataset $\mathcal{D}_{init}$, and the output is used in the inference stage; afterward, once a request to delete the data $\mathcal{D}_{m}$ is received, the updated model $\textit{W}^u$ can be obtained through the unlearning stage, when the data set becomes $\mathcal{D} \setminus \mathcal{D}_{m}$. The process pointed by the red arrow is to apply the updated model $\textit{W}^u$ directly to the inference stage, i.e., approximate unlearning; the process pointed by the green arrow is to start retraining the initial model $\textit{W}^*$ on the new data set $\mathcal{D} \setminus \mathcal{D}_{m}$ from scratch, i.e., exact unlearning \cite{mahadevan2021certifiable}.}
\label{fig3}
\end{figure}

The unlearning problem is defined as a kind of game between two parties, the service provider $\mathcal{S}$, and the user population $\mathcal{U}$ by Bourtoule et al. \cite{bourtoule2021machine}. The service provider $\mathcal{S}$ can be an organization that can collect various users' information, and the collected information is stored in the form of a dataset $\mathcal{D}$. The service provider $\mathcal{S}$ uses this data to train and test a machine learning model $\mathcal{M}$ as a way to provide a intelligent service to the user $\mathcal{U}$. Then according to the GDPR, any user $u$ $\in$ $\mathcal{U}$ has the right to request the removal of part of the data $d_u$ from $\mathcal{D}$, and the service provider $\mathcal{S}$ must execute it. Thus, the service provider $\mathcal{S}$ must modify the model $\mathcal{M}$ to generate a new model $\mathcal{M} \neg {d_u}$, which represents a model without trained data $d_u$.

Guo et al. \cite{guo2019certified} propose a similar concept, \textit{certified removal}, from an accuracy perspective. $\mathcal{D}$ is assumed to be a training dataset and \textit{A} is the learning algorithm used to train $\mathcal{D}$, resulting in model $\textit{h} \in \mathcal{H}$, that is, \textit{A}: $\mathcal{D} \to \mathcal{H}$. When a request is made to remove sample \textit{x} from $\mathcal{D}$, this results in a data removal mechanism \textit{M}, one that can be applied to \textit{A}($\mathcal{D}$) and removes the effects of \textit{x}. If the removal is successful, the output of \textit{M} should be much close to the output of \textit{A} applied on $\mathcal{D} \neg \textit{x}$. Given $\epsilon > 0$, the removal mechanism \textit{M} is said to perform $\epsilon$-\textit{certified removal} for learning algorithm \textit{A} if $\forall \mathcal{T} \subseteq \mathcal{H}$, $\mathcal{D} \subseteq \mathcal{X}$, \textit{x} $\in \mathcal{D}$:
$$ e^{-\epsilon} \leq \frac{\textit{P}(\textit{M}(\textit{A}(\mathcal{D}),\mathcal{D},\textit{x})\in \textit{T})}{\textit{P}(\textit{A}(\mathcal{D}\neg \textit{x})\in \textit{T})} \leq e^\epsilon. $$
The above definition states that the ratio between the likelihood of a model after the removal of sample \textit{x} and a model that was never trained on sample \textit{x} is close to one for all models, all possible data sets, and all removed samples.

However, some researchers have also proposed different views on defining machine unlearning. Thudi et al. \cite{thudi2021necessity} argue that machine unlearning should be divided into \textit{exact unlearning} \cite{ullah2021machine} and \textit{approximate unlearning} \cite{mahadevan2021certifiable}. Exact unlearning means the model outputs after removing the sample \textit{x} is the same as the one that was never trained on the removed sample \textit{x}; approximate unlearning means the model and dataset are adjusted so that it does not need to be retrained from scratch. Current definitions of machine unlearning seek to make the output of approximate unlearning as close as possible to the output of exact unlearning. They suggest this definition is incorrect because the same model can be obtained even when trained on a different data set. Moreover, this definition only applies at the algorithmic level.

Figure 5 illustrates how machine unlearning algorithms can be applied to machine learning models and the essential difference between \textbf{exact unlearning} and \textbf{approximate unlearning} by defining a typical machine learning pipeline \cite{mahadevan2021certifiable} with the three phases of model training, inference and data unlearning. Finally, we summarize reviewed studies relatively to exact and approximate unlearning, as shown in Table 1.

\begin{table}[h]
\begin{center}
\begin{minipage}{\textwidth}
\caption{Summary of reviewed studies relatively to exact and approximate unlearning.}\label{tab2}
\begin{tabular*}{\textwidth}{@{\extracolsep{\fill}}lccclc@{\extracolsep{\fill}}}
\toprule%
&& \multicolumn{2}{@{}c@{}}{Unlearning type} && \\\cmidrule{3-4}%
&& Exact & Approximate &&\\
Author & Year & unlearning & unlearning & \multicolumn{1}{c}{Approach} & Ref. \\
\midrule
\multirow{2}*{Cao \& Yang} & \multirow{2}*{2015} & \multirow{2}*{\checkmark} && Summations following SQ& \multirow{2}*{\cite{cao2015towards}}\\
&&&& learning &\\
Cao et al. & 2018 & \checkmark && Causal unlearning & \cite{cao2018efficient}\\
Ullah et al. & 2021 & \checkmark && Total variation stability & \cite{ullah2021machine}\\
Kashef & 2021 & \checkmark && Decremental unlearning & \cite{kashef2021boosted}\\
Schelter & 2021 & \checkmark && Incremental maintenance & \cite{scheltertowards}\\
Jose \& Simeone & 2021 & \checkmark && Pac-bayesian & \cite{jose2021unified}\\
Bourtoule & 2021 & \checkmark && Federated learning (SISA) & \cite{bourtoule2021machine}\\
Liu et al. & 2021 & \checkmark && Federated unlearning & \cite{liu2021federaser}\\
Brophy \& Lowd & 2021 & \checkmark && Random forests & \cite{brophy2021machine}\\
Wu et al. & 2022 & \checkmark && Federated unlearning & \cite{wu2022federated}\\
Guo et al. & 2019 && \checkmark & Newton method & \cite{guo2019certified}\\
\multirow{2}*{Du et al.} & \multirow{2}*{2019} && \multirow{2}*{\checkmark} & Exploding loss and & \multirow{2}*{\cite{du2019lifelong}}\\
&&&& catastrophic forgetting &\\
Baumhauer et al. & 2020 && \checkmark & Linear filtration & \cite{baumhauer2020machine}\\
Golatkar et al. & 2020 && \checkmark & Differential privacy & \cite{golatkar2020eternal}\\
\multirow{2}*{Wu et al.} & \multirow{2}*{2020} && \multirow{2}*{\checkmark} & Rapid retraining by storing & \multirow{2}*{\cite{wu2020deltagrad}}\\
&&&& training data &\\
Graves et al. & 2020 && \checkmark & Amnesiac unlearning & \cite{graves2020amnesiac}\\
Golatkar et al. & 2021 && \checkmark & Mixed-privacy setting & \cite{golatkar2021mixed}\\
Izzo et al. & 2021 && \checkmark & Influence method & \cite{izzo2021approximate}\\
Neel et al. & 2021 && \checkmark & Gradient-based method & \cite{neel2021descent}\\
Thudi et al. & 2021 && \checkmark & Verification unlearn-error & \cite{thudi2021unrolling}\\
Warnecke et al. & 2021 && \checkmark & Parameters updates & \cite{warnecke2021machine}\\
He et al. &2021 && \checkmark & Intermeidate models & \cite{he2021deepobliviate}\\
\multirow{2}*{Gong et al.} & \multirow{2}*{2021} && \multirow{2}*{\checkmark} & Particle-Based Bayesian & \multirow{2}*{\cite{gong2021forget}}\\
&&&& Federated Unlearning &\\
Guo et al. & 2022 && \checkmark & Vertical unlearning & \cite{guo2022vertical}\\

\botrule
\end{tabular*}
\end{minipage}
\end{center}
\end{table}

\subsection{Exact Unlearning}
\textbf{Exact unlearning} \cite{ullah2021machine} means that in the case of direct use of user data to build a machine learning model, such as a prediction task, a reasonable criterion is that the state of the system is adjusted to what it would be in the complete absence of user data.

Ullah et al. proposed an efficient machine unlearning algorithm, \textbf{total variation stability}, for the convex risk minimization problem, provided that the following three properties are satisfied.
\begin{itemize}
    \item in the stream, at every time point, the output model should be indistinguishable from what we would have obtained if trained on the updated dataset;
    \item the run-time of unlearning method should be small;
    \item the output model should be effective in terms of the accuracy.
\end{itemize}

There exist several exact unlearning approaches, for example, in support vector machines \cite{cauwenberghs2000incremental, tsai2014incremental, karasuyama2010multiple,kashef2021boosted}, naive bayes \cite{cao2015towards, scheltertowards,jose2021unified}, collaborative filtering and ridge regression. This subsection will introduce and analyze several representative \textit{exact unlearning} approaches.

\subsubsection{Machine Unlearning's First Proposed}
Cao and Yang first introduced the concept of machine unlearning in \cite{cao2015towards}. They present an unlearning method by transforming the model learning algorithm into a summation form that follows the statistical query (SQ) learning \cite{kearns1998efficient}. The unlearning method is performed by simply updating a small number of summations from the training dataset. The small number of summations is set in a layer between the machine learning algorithm and the model's training data to break down the dependencies. The learning algorithm only depends on the summations.

The authors implemented the unlearning method based on non-adaptive SQ learning (i.e., all SQs are determined upfront before the algorithm starts) and adaptive SQ learning (i.e., the later SQs may depend on the earlier SQ results). In this case, their summation forms can be implemented in many machine learning models and all stages.

\subsubsection{\textbf{SISA} Training Approach}

Bourtoule et al. \cite{bourtoule2021machine} introduce \textbf{SISA} training approach, short for Sharded, Isolated, Sliced, and Aggregated training. This framework expedites the unlearning process by strategically limiting the influence of a data point in the training procedure, which is illustrated in Figure 6.

In this model, the authors slice the original dataset $\mathcal{D}$ into ${s}$ sub-datasets $\mathcal{D}_{1}$ to $\mathcal{D}_{s}$, and the machine learning network into ${s}$ sub-networks $\mathcal{M}_{1}$ to $\mathcal{M}_{s}$. Each sub-dataset  $\mathcal{D}_{k}$ is trained by the corresponding sub-network $\mathcal{M}_{k}$, and the final training results are integrated by the aggregation algorithm. In this series, if a portion of data is requested to be deleted, it is only necessary to remove it from the sub-dataset and retrain it. Finally, the training results are reintegrated to obtain the new training results. This approach reduces the unnecessary data and model training process and dramatically reduces the time and computational power consumed by machine retraining.

\begin{figure}[h]
\centerline{\includegraphics[scale=0.5]{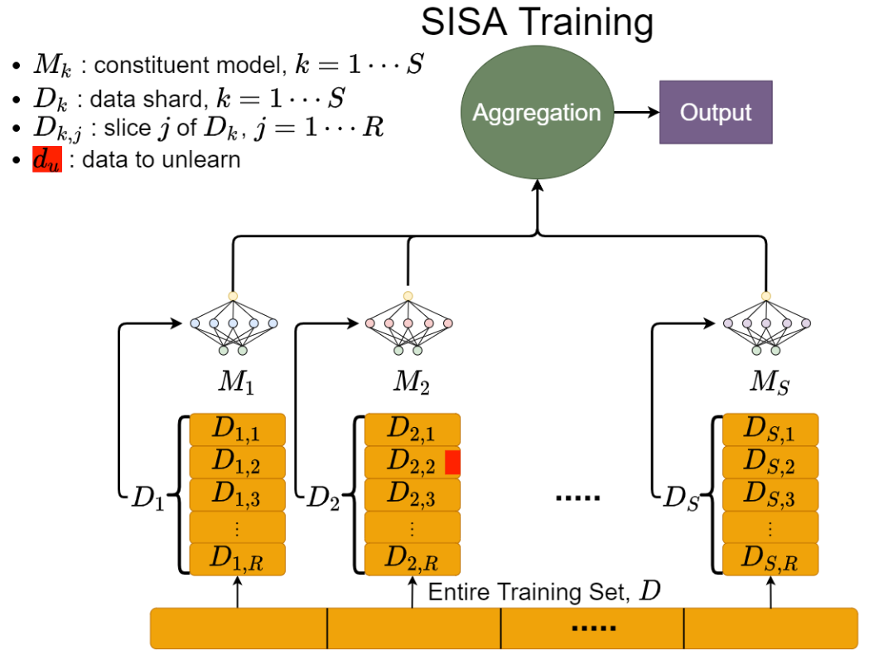}}
\caption{The \textbf{SISA} approach is presented in the form of federated learning. In the \textbf{SISA} structure, separated machine learning models are trained on separated data blocks and the outputs are aggregated in the final inference stage. \cite{bourtoule2021machine}.}
\label{fig4}
\end{figure}

In the paper, the authors illustrate that for simple learning tasks, the \textbf{SISA} training approach can accomplish the unlearning requests quickly without affecting the accuracy of the model. However, for complex learning tasks, the \textbf{SISA} training approach needs to be combined with other learning methods, such as transfer learning, to reduce the impact on model accuracy and to complete the unlearning requests quickly.

\subsection{Approximate Unlearning}
\textbf{Approximate unlearning} is a method for approximating the effect of model retraining by adjusting machine learning models and data sets. Mahadevan et al. \cite{mahadevan2021certifiable} summarize that approximate unlearning methods can be roughly divided into three groups, and this subsection will introduce several methods on this basis.

\subsubsection{First Group}
The first group updates the machine learning model by retraining it with the remaining data and injecting optimal noise based on the principle of \textbf{Fisher information matrix} \cite{martens2014new} to control the certifiability.

\textbf{Differential privacy} \cite{dwork2014algorithmic,chaudhuri2008privacy} can guarantee that the parameters of the trained model do not leak any individual information. Golatkar et al. \cite{golatkar2020eternal,golatkar2020forgetting} proposed a method to selectively unlearn the dataset and update machine learning models based on differential privacy methods. They propose a method for "scrubbing" the weights to remove specific training data used to train deep neural networks. This method does not require retraining from scratch or accessing the data initially used for training. Instead, this method modifies the weights of the model so that any probe function of the weights approximates the same function as the weights of a network that has not been trained on these particular data.

In \cite{golatkar2021mixed}, Golatkar et al. introduce a new concept for machine unlearning, mixed-privacy setting, based on their previous research. According to this method, a "core" subset of the training samples need not be unlearned. Similar to \cite{golatkar2020eternal,golatkar2020forgetting}, this method allows to effectively remove all the information contained in the non-core data by simply setting a subset of the weights to zero with minimal performance loss. They demonstrate that this method yields significant improvements of unlearning in accuracy and guarantees for a large-scale vision classification task.

\subsubsection{Second Group}
The second group updates the machine learning model with the deleted data during the unlearning, they perform a \textbf{Newton's method} \cite{koh2017understanding} to estimate the impact of the deleted data on the model and removing it. The work \cite{guo2019certified} attempted approximate retraining by taking a single Newton’s step. This can be formed as 
$$\theta_{Newton} = \theta^{full} - [\nabla_{\theta}^2 \textit{L}^{\backslash k} (\theta^{full})]^{-1}\nabla_\theta \textit{L}^{\backslash k}(\theta^{full}).$$ 
Where $\theta \in \mathbb{R}^d$ denotes the model parameters, \textit{k} denotes the number of data points to be deleted from the model, $\theta^{full}$ = argmin $\theta \textit{L}^{full}(\theta)$ are the model parameters when fitted to the full dataset, $\textit{L}^{\backslash k}(\theta)$ is the loss on the LKO dataset. When the loss function is quadratic in $\theta$, the approximation to
$\textit{L}^{\backslash k}$ is just $\textit{L}^{\backslash k}$ itself which means Newton’s method is effective for solving this issue.

Izzo et al. \cite{izzo2021approximate} introduces the \textbf{Influence method} \cite{giordano2019swiss,koh2017understanding} to estimate the influence of a particular training point on the model’s predictions. The influence method can be formed under suitable assumptions on the loss function \textit{l}: $\theta(w)\equiv$ argmin$_\theta \sum_{i=1}^n w_i \textit{l} (x_i, y_i;\theta),$ where \textit{n} denotes the total number of training points, $\textit{X} = [x_i, ..., x_n]^\top \in \mathbb{R}^{n \times d}$ is the data matrix for full set of training data $\mathcal{D}^{full}$, $\textit{Y} = [y_i, ..., y_n]^\top \in \mathbb{R}^n$ is the response vector for $\mathcal{D}^{full}$, \textit{d} denotes the data dimension. In this setting, $\theta^{full} = \theta(1)$ where 1 is the all 1s vector and $\theta^{\backslash k} = \theta((\underbrace{\theta, ...,}_{k} \underbrace{1, ...}_{n-k})^\top) .$ The influence function approach uses the linear approximation \cite{giordano2019swiss,koh2017understanding}:
$$\theta_{inf} = \theta^{full} - [\nabla_{\theta}^2 \textit{L}^{full} (\theta^{full})]^{-1}\nabla_\theta \textit{L}^{\backslash k}(\theta^{full}).$$
to $\theta(w)$ about $w = 1$ to estimate $\theta^{\backslash k}.$ Therefore, they propose a unlearning method based on the influence method principle that the computational cost is linearly related to the feature dimension \textit{d}, i.e., $O(d^2)$, and is independent of the number of training data \textit{n}. And this method is applicable to both linear regression and logistic regression models.

The influence method explains the principle of machine unlearning at a higher level. In other words, the influence method-based unlearning can compute the impact of the deleted data relatively to the parameters of the trained model for removing the influence and updates the parameters.

\subsubsection{Third Group}
The third group stores the data and related information during the machine learning model training and use them to update the model when a request to delete the data is made \cite{wu2020deltagrad,neel2021descent}.

Graves et al. \cite{graves2020amnesiac} proposed the concept of \textbf{Amnesiac Unlearning}, where the model owner stores the sensitive data and parameters in the form of batches during the training process. When a request for deleting the data is made, the model owner does not perform the parameter update of the batches containing the deleted data. This process can also be interpreted as selectively undoing specific machine learning steps containing sensitive data. The model training can be regarded as a series of parameter updates to the initial model parameters. The model parameters can be expressed as:
$$\theta_M = \theta_{initial} + \sum_{e=1}^{E} \sum_{b=1}^{B} \Delta_{\theta_{e,b}}$$
where $\theta_{initial}$ is the initial model parameters, model $M$ is trained for $E$ epochs, each epoch consists of $B$ batches. The model parameters are updated after each batch by an amount $\Delta_{\theta_{e,b}}.$ During training, the model owner stores a list $SB$, which refers to the batches contain the sensitive data. When the request of removing data  $s$ belonging to the batch $b$, where $sb \in SB$, is received, the amnesiac unlearning method can simply remove the parameter updates from the learning parameters $\theta_M$ to get the $\theta_{M'}$:
$$\theta_{M'} = \theta_{initial} + \sum_{e=1}^{E} \sum_{b=1}^{B} \Delta_{\theta_{e,b}} - \sum_{sb=1}^{SB} \Delta_{\theta_{sb}} = \theta_M - \sum_{sb=1}^{SB} \Delta_{\theta_{sb}}.$$

This approach has a potential drawback in that the model owner needs a large amount of storage space for storing sensitive data and related parameters. However, the authors argue that this space cost is much less than the computational and time cost of the exact unlearning methods.

\subsubsection{Evaluation Metrics}
For approximate unlearning, in addition to designing an effective and fast algorithm for data deletion, it is a significant challenge to evaluate the quality of an approximate unlearning method properly. As a result, many researchers proposed effective evaluation metrics for their algorithms.

In \cite{mahadevan2021certifiable}, the authors defined three evaluation metrics to measure the performance of different unlearning methods in terms of effectiveness, certifiability and efficiency on the basis of the \textit{Symmetric Absolute Percentage Error} (SAPE) defined as: $$SAPE(a, b) = \frac{\mid b - a \mid}{\mid b \mid + \mid a \mid} \cdot 100\%. $$ 
\begin{itemize}
    \item \textbf{Effectiveness} is used to measure the prediction accuracy of a machine learning model. The error in test accuracy $Acc_{Err}$ of the updated model $w^u$ is defined as: $$Acc_{Err} = SAPE(Acc_{test}^*, Acc_{test}^u)$$
    where $Acc_{test}^u$ denotes the accuracy of the updated model $w^u$ on the test dataset $\mathcal{D}_{test}$, $Acc_{test}^*$ denotes the optimal accuracy of the regression model on the same dataset. The lower value of $Acc_{Err}$ means that the prediction accuracy of $w^u$ is closer to the accuracy of the initial model (in which the noise value $\sigma = 0$), i.e., $w^u$ is more effective.
    
    \item \textbf{Certifiability} is used to measure how well the updated model $w^u$ has unlearned the delated data. For the certifiability, both the updated model and the fully retrained model are considered, the disparity in accuracy of the two models $Acc_{Dis}$ is defined as: $$Acc_{Dis} = SAPE(Acc_{del}^*, Acc_{del}^u)$$
    where $Acc_{del}^u$ denotes the accuracy on the deleted data $\mathcal{D}_{del}$ for the updated model $w^u$, and $Acc_{del}^*$ denotes the accuracy on the deleted data for the fully retrained model. The lower value of $Acc_{Dis}$ means that the updated model is more similar to the fully retrained model, i.e. the updated model $w^u$ has higher certifiability.
    
    \item \textbf{Efficiency} is used to measure the speed-up performance of running the algorithm to obtain the updated model $w^u$ and the fully retrained model $w^*$: $$ speed-up = \frac{{time}\ {taken}\ {to}\ {obtain}\ w^*}{{time}\ {taken}\ {to}\ {obtain}\ w^u} \cdot x.$$
\end{itemize}

Izzo et al. \cite{izzo2021approximate} introduced two metrics, \textbf{$L^2$ distance} and \textbf{Feature injection test}, to evaluate the effectiveness of an approximate data deletion method. 
\begin{itemize}
    \item \textbf{$L^2$ distance} between the updated model and the fully retrained model is a relatively common method used to measure the accuracy of approximate unlearning. A lower value of $L^2$ distance indicates that the predictive ability of the updated model is closer to that of the fully retrained model.
    
    \item \textbf{Feature injection test} is injected as a strong signal (an extra feature) into the remaining dataset, which the model (updated and fully retrained model) expects to learn. The authors measure the effectiveness of the approximate deletion method by observing the performance of the model's learned parameters before and after the removal of this particular feature.

\end{itemize}

\section{Discussion on Data Lineage}

\begin{figure*}[!htpb]
\centerline{\includegraphics[scale=0.32]{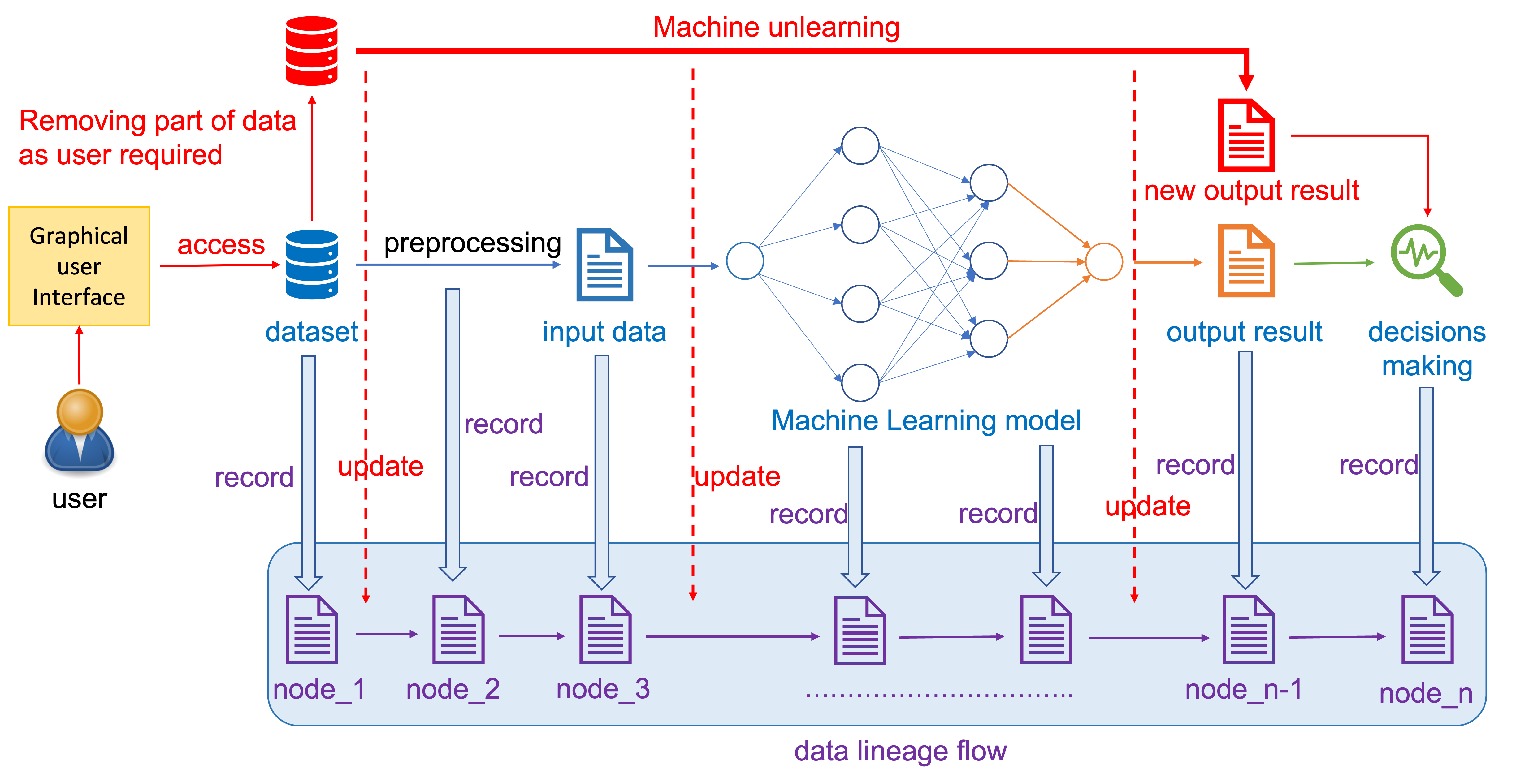}}
\caption{Data lineage management for machine learning data flows recording and machine unlearning updating. For each step in the machine learning system, from the original dataset to optimizing each parameter in the training process to getting the final training results and analyzing the results, all data and changes will be recorded in the data lineage management system. This series of records of data flow characteristics and changes allow developers to control and track any subtle differences in the model learning process at any time.}
\label{fig1}
\end{figure*}

In the process of protecting the privacy of machine learning models, the tracking of data flow is an essential part of the process \cite{zhang2017diagnosing,baracaldo2017mitigating,baracaldo2018detecting,luo2021roadmap,thiago2020managing}. Therefore, this section discusses the role of data lineage management techniques in the privacy protection of machine learning models.

Data lineage tracks data movement over time from the source system to different forms of persistence and transformations and ultimately to data consumption by an application or analytics model. The data lineage management system can monitor any data changes in the machine learning model that occur at any point in time \cite{li2022making}. Therefore, the combination with the data lineage management system can effectively enhance the security protection of machine learning models. 

Data lineage management can be applied to defend against particular cyberattacks, such as data poisoning attacks, which can be viewed as integrity attacks. The attacker affects the model's prediction of the correct output by tampering with the training data. Even the attacker's goal is to have their input accepted as the model's training data.

Baracaldo et al. \cite{baracaldo2017mitigating} proposed to identify poisonous data by using the sytem's lineage about the sources, transformation and destinations of data points in the training dataset of a machine learning model as part of a filtering algorithm, which is also known as a method for detecting causative attacks. With this approach, online and periodically retrained machine learning systems can discriminate between data sources in a potentially adversarial environment. Subsequently, they applied this approach to identify poisonous data injection in the Internet of Things environment as well \cite{baracaldo2018detecting}.

The Tensorflow team of Google developed a version control platform for machine learning data lineage management, Machine Learning Metadata (MLMD). MLMD can be viewed as a library to track the complete data lineage of the entire machine learning workflow, including the metadata, data preprocessing, feature selection, model training, prediction, evaluation, deployment and so on. This work aims to answer questions like,

\begin{itemize}
    \item What hyperparameters were this model used?
    \item Which dataset was this model trained on?
    \item Which version of libraries were used to build this model?
    \item Which pipeline was used to build this model?
    \item Which version of this model was last deployed?
    \item What is the reason for this model's failure?
\end{itemize}

MLMD can be implemented in various machine learning pipelines to record and control all the data generated during model training. It can help developers analyze all model data transformations, including parameter updates and debugging of errors. Furthermore, from the security perspective, this metadata platform also provides ideas for the research of combining machine unlearning with data lineage.

Figure 7 explains how the data lineage management technique works in a machine learning system. With the machine unlearning approach, data lineage can still play an important role. For example, developers use machine unlearning when a user wants to withdraw sensitive personal information used to train a machine learning model or when a malicious data injection attack is detected. A part of the training data needs to be removed from the dataset to retrain the model. This process is also recorded in the data lineage management system without reservation.

\section{Challenges}
This paper describes the security risks and privacy protection issues associated with machine learning models. Both machine unlearning and data lineage management systems can play a role in addressing these issues. However, research in this area is just beginning, and researchers still face many challenges. For example, how machine unlearning can efficiently handle large amounts of data deletion tasks in a big data environment; how to quickly respond to privacy theft when machine learning platforms encounter it; and how data lineage management systems can make the most of the privacy-preserving aspects of machine learning.

\subsection{Machine Unlearning Algorithms}
There are not many machine unlearning algorithms designed for the privacy preservation of machine learning models. Instead, algorithms with high adaptability are necessary for different user needs or data diversity. For example, the superiority of the SISA algorithm can be demonstrated when the amount of data requested for deletion is small. However, when the amount of data requested for deletion is large, the retraining approach becomes more applicable.

\subsection{Active and Passive Unlearning}
The machine unlearning methods we are discussing are all based on the active unlearning at the will of the data holder. However, passive unlearning is also a good option for the CIA property of machine learning models.  When an attacker performs a CIA attack on a machine learning model, the data holder or the machine learning platform does not discover this attacker's behavior in time, which leads to the private data being compromised before taking countermeasures (such as machine unlearning methods). In this case, the passive model unlearning method can delete the data in time when the machine learning system is attacked, thus minimizing the loss of the data holder.

\subsection{Privacy Risks of Machine Unlearning}
The original intent of machine unlearning was to prevent privacy leaks caused by machine learning. However, some researchers have questioned the privacy-preserving effects of machine unlearning in recent years. Chen et al. \cite{chen2021machine} propose that machine unlearning methods can also be attacked and leak the privacy of models in specific scenarios, such as membership inference attacks \cite{shokri2017membership, yeom2018privacy, sablayrolles2019white, hayes2019logan}. They designed novel membership attacks and conducted experimental evaluations against two machine unlearning approaches, retraining the machine learning model from scratch and the SISA approach. The experimental results show that their attack methods significantly impact unlearning methods that handle tedious tasks, i.e., retraining from scratch. In contrast, they have less impact on distributed unlearning models like SISA.

\subsection{Working with Data Lineage}
At the level of security and privacy protection for machine learning, the data lineage management system can trace all data and changes in the model. With the introduction of the machine unlearning approach as a protection mechanism for machine learning models, it is imperative to use it in conjunction with a data lineage management system. The machine unlearning approach can be considered another model independent of the machine learning model used for training in the same environment. Any data changes in the machine unlearning model directly affect the security of the machine learning model used for training and, therefore, should be recorded by the data lineage management system.

\section{Conclusion}
This paper starts with the right to be forgotten of the GDPR regulations. Then, it discusses the security concerns in machine learning models and the possible privacy breaches to the data holders used for training. In this process, machine unlearning methods and data lineage management play an essential role in machine learning privacy protection. Furthermore, the challenges this research area may encounter in the future are elaborated. More and more machine learning models appear in our lives at a swift pace. While we enjoy the convenience of technological development, we cannot let down our guard on the potential security threats that may exist.

\bmhead{Acknowledgments}

This research was partially supported by the Japan Science and Technology Agency (JST) Strategic International Collaborative Research Program (SICORP).

The first author was supported by JST SPRING, Grant Number JPMJSP2136.

\bmhead{Conflict of Interest}
On behalf of all the authors, the corresponding author states that there is no conflict of interest.

\bibliography{bibdatabase}


\begin{thebibliography}{60}
\ifx \bisbn   \undefined \def \bisbn  #1{ISBN #1}\fi
\ifx \binits  \undefined \def \binits#1{#1}\fi
\ifx \bauthor  \undefined \def \bauthor#1{#1}\fi
\ifx \batitle  \undefined \def \batitle#1{#1}\fi
\ifx \bjtitle  \undefined \def \bjtitle#1{#1}\fi
\ifx \bvolume  \undefined \def \bvolume#1{\textbf{#1}}\fi
\ifx \byear  \undefined \def \byear#1{#1}\fi
\ifx \bissue  \undefined \def \bissue#1{#1}\fi
\ifx \bfpage  \undefined \def \bfpage#1{#1}\fi
\ifx \blpage  \undefined \def \blpage #1{#1}\fi
\ifx \burl  \undefined \def \burl#1{\textsf{#1}}\fi
\ifx \doiurl  \undefined \def \doiurl#1{\url{https://doi.org/#1}}\fi
\ifx \betal  \undefined \def \betal{\textit{et al.}}\fi
\ifx \binstitute  \undefined \def \binstitute#1{#1}\fi
\ifx \binstitutionaled  \undefined \def \binstitutionaled#1{#1}\fi
\ifx \bctitle  \undefined \def \bctitle#1{#1}\fi
\ifx \beditor  \undefined \def \beditor#1{#1}\fi
\ifx \bpublisher  \undefined \def \bpublisher#1{#1}\fi
\ifx \bbtitle  \undefined \def \bbtitle#1{#1}\fi
\ifx \bedition  \undefined \def \bedition#1{#1}\fi
\ifx \bseriesno  \undefined \def \bseriesno#1{#1}\fi
\ifx \blocation  \undefined \def \blocation#1{#1}\fi
\ifx \bsertitle  \undefined \def \bsertitle#1{#1}\fi
\ifx \bsnm \undefined \def \bsnm#1{#1}\fi
\ifx \bsuffix \undefined \def \bsuffix#1{#1}\fi
\ifx \bparticle \undefined \def \bparticle#1{#1}\fi
\ifx \barticle \undefined \def \barticle#1{#1}\fi
\bibcommenthead
\ifx \bconfdate \undefined \def \bconfdate #1{#1}\fi
\ifx \botherref \undefined \def \botherref #1{#1}\fi
\ifx \url \undefined \def \url#1{\textsf{#1}}\fi
\ifx \bchapter \undefined \def \bchapter#1{#1}\fi
\ifx \bbook \undefined \def \bbook#1{#1}\fi
\ifx \bcomment \undefined \def \bcomment#1{#1}\fi
\ifx \oauthor \undefined \def \oauthor#1{#1}\fi
\ifx \citeauthoryear \undefined \def \citeauthoryear#1{#1}\fi
\ifx \endbibitem  \undefined \def \endbibitem {}\fi
\ifx \bconflocation  \undefined \def \bconflocation#1{#1}\fi
\ifx \arxivurl  \undefined \def \arxivurl#1{\textsf{#1}}\fi
\csname PreBibitemsHook\endcsname

\bibitem{baracaldo2017mitigating}
\begin{bchapter}
\bauthor{\bsnm{Baracaldo}, \binits{N.}},
\bauthor{\bsnm{Chen}, \binits{B.}},
\bauthor{\bsnm{Ludwig}, \binits{H.}},
\bauthor{\bsnm{Safavi}, \binits{J.A.}}:
\bctitle{Mitigating poisoning attacks on machine learning models: A data
  provenance based approach}.
In: \bbtitle{Proceedings of the 10th ACM Workshop on Artificial Intelligence
  and Security},
pp. \bfpage{103}--\blpage{110}
(\byear{2017})
\end{bchapter}
\endbibitem

\bibitem{liu2022backdoor}
\begin{botherref}
\oauthor{\bsnm{Liu}, \binits{Y.}},
\oauthor{\bsnm{Fan}, \binits{M.}},
\oauthor{\bsnm{Chen}, \binits{C.}},
\oauthor{\bsnm{Liu}, \binits{X.}},
\oauthor{\bsnm{Ma}, \binits{Z.}},
\oauthor{\bsnm{Wang}, \binits{L.}},
\oauthor{\bsnm{Ma}, \binits{J.}}:
Backdoor defense with machine unlearning.
arXiv preprint arXiv:2201.09538
(2022)
\end{botherref}
\endbibitem

\bibitem{bourtoule2021machine}
\begin{bchapter}
\bauthor{\bsnm{Bourtoule}, \binits{L.}},
\bauthor{\bsnm{Chandrasekaran}, \binits{V.}},
\bauthor{\bsnm{Choquette-Choo}, \binits{C.A.}},
\bauthor{\bsnm{Jia}, \binits{H.}},
\bauthor{\bsnm{Travers}, \binits{A.}},
\bauthor{\bsnm{Zhang}, \binits{B.}},
\bauthor{\bsnm{Lie}, \binits{D.}},
\bauthor{\bsnm{Papernot}, \binits{N.}}:
\bctitle{Machine unlearning}.
In: \bbtitle{2021 IEEE Symposium on Security and Privacy (SP)},
pp. \bfpage{141}--\blpage{159}
(\byear{2021}).
\bcomment{IEEE}
\end{bchapter}
\endbibitem

\bibitem{al2019privacy}
\begin{barticle}
\bauthor{\bsnm{Al-Rubaie}, \binits{M.}},
\bauthor{\bsnm{Chang}, \binits{J.M.}}:
\batitle{Privacy-preserving machine learning: Threats and solutions}.
\bjtitle{IEEE Security \& Privacy}
\bvolume{17}(\bissue{2}),
\bfpage{49}--\blpage{58}
(\byear{2019})
\end{barticle}
\endbibitem

\bibitem{scheltertowards}
\begin{botherref}
\oauthor{\bsnm{Schelter}, \binits{S.}}:
Towards efficient machine unlearning via incremental view maintenance
\end{botherref}
\endbibitem

\bibitem{graves2020amnesiac}
\begin{botherref}
\oauthor{\bsnm{Graves}, \binits{L.}},
\oauthor{\bsnm{Nagisetty}, \binits{V.}},
\oauthor{\bsnm{Ganesh}, \binits{V.}}:
Amnesiac machine learning.
arXiv preprint arXiv:2010.10981
(2020)
\end{botherref}
\endbibitem

\bibitem{chen2021machine}
\begin{bchapter}
\bauthor{\bsnm{Chen}, \binits{M.}},
\bauthor{\bsnm{Zhang}, \binits{Z.}},
\bauthor{\bsnm{Wang}, \binits{T.}},
\bauthor{\bsnm{Backes}, \binits{M.}},
\bauthor{\bsnm{Humbert}, \binits{M.}},
\bauthor{\bsnm{Zhang}, \binits{Y.}}:
\bctitle{When machine unlearning jeopardizes privacy}.
In: \bbtitle{Proceedings of the 2021 ACM SIGSAC Conference on Computer and
  Communications Security},
pp. \bfpage{896}--\blpage{911}
(\byear{2021})
\end{bchapter}
\endbibitem

\bibitem{gao2022deletion}
\begin{botherref}
\oauthor{\bsnm{Gao}, \binits{J.}},
\oauthor{\bsnm{Garg}, \binits{S.}},
\oauthor{\bsnm{Mahmoody}, \binits{M.}},
\oauthor{\bsnm{Vasudevan}, \binits{P.N.}}:
Deletion inference, reconstruction, and compliance in machine (un) learning.
arXiv preprint arXiv:2202.03460
(2022)
\end{botherref}
\endbibitem

\bibitem{marchant2021hard}
\begin{botherref}
\oauthor{\bsnm{Marchant}, \binits{N.G.}},
\oauthor{\bsnm{Rubinstein}, \binits{B.I.}},
\oauthor{\bsnm{Alfeld}, \binits{S.}}:
Hard to forget: Poisoning attacks on certified machine unlearning.
arXiv preprint arXiv:2109.08266
(2021)
\end{botherref}
\endbibitem

\bibitem{baracaldo2018detecting}
\begin{bchapter}
\bauthor{\bsnm{Baracaldo}, \binits{N.}},
\bauthor{\bsnm{Chen}, \binits{B.}},
\bauthor{\bsnm{Ludwig}, \binits{H.}},
\bauthor{\bsnm{Safavi}, \binits{A.}},
\bauthor{\bsnm{Zhang}, \binits{R.}}:
\bctitle{Detecting poisoning attacks on machine learning in iot environments}.
In: \bbtitle{2018 IEEE International Congress on Internet of Things (ICIOT)},
pp. \bfpage{57}--\blpage{64}
(\byear{2018}).
\bcomment{IEEE}
\end{bchapter}
\endbibitem

\bibitem{chundawat2022zero}
\begin{botherref}
\oauthor{\bsnm{Chundawat}, \binits{V.S.}},
\oauthor{\bsnm{Tarun}, \binits{A.K.}},
\oauthor{\bsnm{Mandal}, \binits{M.}},
\oauthor{\bsnm{Kankanhalli}, \binits{M.}}:
Zero-shot machine unlearning.
arXiv preprint arXiv:2201.05629
(2022)
\end{botherref}
\endbibitem

\bibitem{toreini2020relationship}
\begin{bchapter}
\bauthor{\bsnm{Toreini}, \binits{E.}},
\bauthor{\bsnm{Aitken}, \binits{M.}},
\bauthor{\bsnm{Coopamootoo}, \binits{K.}},
\bauthor{\bsnm{Elliott}, \binits{K.}},
\bauthor{\bsnm{Zelaya}, \binits{C.G.}},
\bauthor{\bsnm{Van~Moorsel}, \binits{A.}}:
\bctitle{The relationship between trust in ai and trustworthy machine learning
  technologies}.
In: \bbtitle{Proceedings of the 2020 Conference on Fairness, Accountability,
  and Transparency},
pp. \bfpage{272}--\blpage{283}
(\byear{2020})
\end{bchapter}
\endbibitem

\bibitem{surma2020hacking}
\begin{bchapter}
\bauthor{\bsnm{Surma}, \binits{J.}}:
\bctitle{Hacking machine learning: towards the comprehensive taxonomy of
  attacks against machine learning systems}.
In: \bbtitle{Proceedings of the 2020 the 4th International Conference on
  Innovation in Artificial Intelligence},
pp. \bfpage{1}--\blpage{4}
(\byear{2020})
\end{bchapter}
\endbibitem

\bibitem{tramer2016stealing}
\begin{bchapter}
\bauthor{\bsnm{Tram{\`e}r}, \binits{F.}},
\bauthor{\bsnm{Zhang}, \binits{F.}},
\bauthor{\bsnm{Juels}, \binits{A.}},
\bauthor{\bsnm{Reiter}, \binits{M.K.}},
\bauthor{\bsnm{Ristenpart}, \binits{T.}}:
\bctitle{Stealing machine learning models via prediction apis}.
In: \bbtitle{25th $\{$USENIX$\}$ Security Symposium ($\{$USENIX$\}$ Security
  16)},
pp. \bfpage{601}--\blpage{618}
(\byear{2016})
\end{bchapter}
\endbibitem

\bibitem{song2017machine}
\begin{bchapter}
\bauthor{\bsnm{Song}, \binits{C.}},
\bauthor{\bsnm{Ristenpart}, \binits{T.}},
\bauthor{\bsnm{Shmatikov}, \binits{V.}}:
\bctitle{Machine learning models that remember too much}.
In: \bbtitle{Proceedings of the 2017 ACM SIGSAC Conference on Computer and
  Communications Security},
pp. \bfpage{587}--\blpage{601}
(\byear{2017})
\end{bchapter}
\endbibitem

\bibitem{shen2016auror}
\begin{bchapter}
\bauthor{\bsnm{Shen}, \binits{S.}},
\bauthor{\bsnm{Tople}, \binits{S.}},
\bauthor{\bsnm{Saxena}, \binits{P.}}:
\bctitle{Auror: Defending against poisoning attacks in collaborative deep
  learning systems}.
In: \bbtitle{Proceedings of the 32nd Annual Conference on Computer Security
  Applications},
pp. \bfpage{508}--\blpage{519}
(\byear{2016})
\end{bchapter}
\endbibitem

\bibitem{alsdurf2020covi}
\begin{botherref}
\oauthor{\bsnm{Alsdurf}, \binits{H.}},
\oauthor{\bsnm{Belliveau}, \binits{E.}},
\oauthor{\bsnm{Bengio}, \binits{Y.}},
\oauthor{\bsnm{Deleu}, \binits{T.}},
\oauthor{\bsnm{Gupta}, \binits{P.}},
\oauthor{\bsnm{Ippolito}, \binits{D.}},
\oauthor{\bsnm{Janda}, \binits{R.}},
\oauthor{\bsnm{Jarvie}, \binits{M.}},
\oauthor{\bsnm{Kolody}, \binits{T.}},
\oauthor{\bsnm{Krastev}, \binits{S.}}, et al.:
Covi white paper.
arXiv preprint arXiv:2005.08502
(2020)
\end{botherref}
\endbibitem

\bibitem{ginart2019making}
\begin{botherref}
\oauthor{\bsnm{Ginart}, \binits{A.}},
\oauthor{\bsnm{Guan}, \binits{M.Y.}},
\oauthor{\bsnm{Valiant}, \binits{G.}},
\oauthor{\bsnm{Zou}, \binits{J.}}:
Making ai forget you: Data deletion in machine learning.
arXiv preprint arXiv:1907.05012
(2019)
\end{botherref}
\endbibitem

\bibitem{fredrikson2015model}
\begin{bchapter}
\bauthor{\bsnm{Fredrikson}, \binits{M.}},
\bauthor{\bsnm{Jha}, \binits{S.}},
\bauthor{\bsnm{Ristenpart}, \binits{T.}}:
\bctitle{Model inversion attacks that exploit confidence information and basic
  countermeasures}.
In: \bbtitle{Proceedings of the 22nd ACM SIGSAC Conference on Computer and
  Communications Security},
pp. \bfpage{1322}--\blpage{1333}
(\byear{2015})
\end{bchapter}
\endbibitem

\bibitem{mahadevan2021certifiable}
\begin{botherref}
\oauthor{\bsnm{Mahadevan}, \binits{A.}},
\oauthor{\bsnm{Mathioudakis}, \binits{M.}}:
Certifiable machine unlearning for linear models.
arXiv preprint arXiv:2106.15093
(2021)
\end{botherref}
\endbibitem

\bibitem{guo2019certified}
\begin{botherref}
\oauthor{\bsnm{Guo}, \binits{C.}},
\oauthor{\bsnm{Goldstein}, \binits{T.}},
\oauthor{\bsnm{Hannun}, \binits{A.}},
\oauthor{\bsnm{Van Der~Maaten}, \binits{L.}}:
Certified data removal from machine learning models.
arXiv preprint arXiv:1911.03030
(2019)
\end{botherref}
\endbibitem

\bibitem{thudi2021necessity}
\begin{botherref}
\oauthor{\bsnm{Thudi}, \binits{A.}},
\oauthor{\bsnm{Jia}, \binits{H.}},
\oauthor{\bsnm{Shumailov}, \binits{I.}},
\oauthor{\bsnm{Papernot}, \binits{N.}}:
On the necessity of auditable algorithmic definitions for machine unlearning.
arXiv preprint arXiv:2110.11891
(2021)
\end{botherref}
\endbibitem

\bibitem{ullah2021machine}
\begin{bchapter}
\bauthor{\bsnm{Ullah}, \binits{E.}},
\bauthor{\bsnm{Mai}, \binits{T.}},
\bauthor{\bsnm{Rao}, \binits{A.}},
\bauthor{\bsnm{Rossi}, \binits{R.A.}},
\bauthor{\bsnm{Arora}, \binits{R.}}:
\bctitle{Machine unlearning via algorithmic stability}.
In: \bbtitle{Conference on Learning Theory},
pp. \bfpage{4126}--\blpage{4142}
(\byear{2021}).
\bcomment{PMLR}
\end{bchapter}
\endbibitem

\bibitem{cao2015towards}
\begin{bchapter}
\bauthor{\bsnm{Cao}, \binits{Y.}},
\bauthor{\bsnm{Yang}, \binits{J.}}:
\bctitle{Towards making systems forget with machine unlearning}.
In: \bbtitle{2015 IEEE Symposium on Security and Privacy},
pp. \bfpage{463}--\blpage{480}
(\byear{2015}).
\bcomment{IEEE}
\end{bchapter}
\endbibitem

\bibitem{cao2018efficient}
\begin{bchapter}
\bauthor{\bsnm{Cao}, \binits{Y.}},
\bauthor{\bsnm{Yu}, \binits{A.F.}},
\bauthor{\bsnm{Aday}, \binits{A.}},
\bauthor{\bsnm{Stahl}, \binits{E.}},
\bauthor{\bsnm{Merwine}, \binits{J.}},
\bauthor{\bsnm{Yang}, \binits{J.}}:
\bctitle{Efficient repair of polluted machine learning systems via causal
  unlearning}.
In: \bbtitle{Proceedings of the 2018 on Asia Conference on Computer and
  Communications Security},
pp. \bfpage{735}--\blpage{747}
(\byear{2018})
\end{bchapter}
\endbibitem

\bibitem{kashef2021boosted}
\begin{barticle}
\bauthor{\bsnm{Kashef}, \binits{R.}}:
\batitle{A boosted svm classifier trained by incremental learning and
  decremental unlearning approach}.
\bjtitle{Expert Systems with Applications}
\bvolume{167},
\bfpage{114154}
(\byear{2021})
\end{barticle}
\endbibitem

\bibitem{jose2021unified}
\begin{bchapter}
\bauthor{\bsnm{Jose}, \binits{S.T.}},
\bauthor{\bsnm{Simeone}, \binits{O.}}:
\bctitle{A unified pac-bayesian framework for machine unlearning via
  information risk minimization}.
In: \bbtitle{2021 IEEE 31st International Workshop on Machine Learning for
  Signal Processing (MLSP)},
pp. \bfpage{1}--\blpage{6}
(\byear{2021}).
\bcomment{IEEE}
\end{bchapter}
\endbibitem

\bibitem{liu2021federaser}
\begin{bchapter}
\bauthor{\bsnm{Liu}, \binits{G.}},
\bauthor{\bsnm{Ma}, \binits{X.}},
\bauthor{\bsnm{Yang}, \binits{Y.}},
\bauthor{\bsnm{Wang}, \binits{C.}},
\bauthor{\bsnm{Liu}, \binits{J.}}:
\bctitle{Federaser: Enabling efficient client-level data removal from federated
  learning models}.
In: \bbtitle{2021 IEEE/ACM 29th International Symposium on Quality of Service
  (IWQOS)},
pp. \bfpage{1}--\blpage{10}
(\byear{2021}).
\bcomment{IEEE}
\end{bchapter}
\endbibitem

\bibitem{brophy2021machine}
\begin{bchapter}
\bauthor{\bsnm{Brophy}, \binits{J.}},
\bauthor{\bsnm{Lowd}, \binits{D.}}:
\bctitle{Machine unlearning for random forests}.
In: \bbtitle{International Conference on Machine Learning},
pp. \bfpage{1092}--\blpage{1104}
(\byear{2021}).
\bcomment{PMLR}
\end{bchapter}
\endbibitem

\bibitem{wu2022federated}
\begin{botherref}
\oauthor{\bsnm{Wu}, \binits{C.}},
\oauthor{\bsnm{Zhu}, \binits{S.}},
\oauthor{\bsnm{Mitra}, \binits{P.}}:
Federated unlearning with knowledge distillation.
arXiv preprint arXiv:2201.09441
(2022)
\end{botherref}
\endbibitem

\bibitem{du2019lifelong}
\begin{bchapter}
\bauthor{\bsnm{Du}, \binits{M.}},
\bauthor{\bsnm{Chen}, \binits{Z.}},
\bauthor{\bsnm{Liu}, \binits{C.}},
\bauthor{\bsnm{Oak}, \binits{R.}},
\bauthor{\bsnm{Song}, \binits{D.}}:
\bctitle{Lifelong anomaly detection through unlearning}.
In: \bbtitle{Proceedings of the 2019 ACM SIGSAC Conference on Computer and
  Communications Security},
pp. \bfpage{1283}--\blpage{1297}
(\byear{2019})
\end{bchapter}
\endbibitem

\bibitem{baumhauer2020machine}
\begin{botherref}
\oauthor{\bsnm{Baumhauer}, \binits{T.}},
\oauthor{\bsnm{Sch{\"o}ttle}, \binits{P.}},
\oauthor{\bsnm{Zeppelzauer}, \binits{M.}}:
Machine unlearning: Linear filtration for logit-based classifiers.
arXiv preprint arXiv:2002.02730
(2020)
\end{botherref}
\endbibitem

\bibitem{golatkar2020eternal}
\begin{bchapter}
\bauthor{\bsnm{Golatkar}, \binits{A.}},
\bauthor{\bsnm{Achille}, \binits{A.}},
\bauthor{\bsnm{Soatto}, \binits{S.}}:
\bctitle{Eternal sunshine of the spotless net: Selective forgetting in deep
  networks}.
In: \bbtitle{Proceedings of the IEEE/CVF Conference on Computer Vision and
  Pattern Recognition},
pp. \bfpage{9304}--\blpage{9312}
(\byear{2020})
\end{bchapter}
\endbibitem

\bibitem{wu2020deltagrad}
\begin{bchapter}
\bauthor{\bsnm{Wu}, \binits{Y.}},
\bauthor{\bsnm{Dobriban}, \binits{E.}},
\bauthor{\bsnm{Davidson}, \binits{S.}}:
\bctitle{Deltagrad: Rapid retraining of machine learning models}.
In: \bbtitle{International Conference on Machine Learning},
pp. \bfpage{10355}--\blpage{10366}
(\byear{2020}).
\bcomment{PMLR}
\end{bchapter}
\endbibitem

\bibitem{golatkar2021mixed}
\begin{bchapter}
\bauthor{\bsnm{Golatkar}, \binits{A.}},
\bauthor{\bsnm{Achille}, \binits{A.}},
\bauthor{\bsnm{Ravichandran}, \binits{A.}},
\bauthor{\bsnm{Polito}, \binits{M.}},
\bauthor{\bsnm{Soatto}, \binits{S.}}:
\bctitle{Mixed-privacy forgetting in deep networks}.
In: \bbtitle{Proceedings of the IEEE/CVF Conference on Computer Vision and
  Pattern Recognition},
pp. \bfpage{792}--\blpage{801}
(\byear{2021})
\end{bchapter}
\endbibitem

\bibitem{izzo2021approximate}
\begin{bchapter}
\bauthor{\bsnm{Izzo}, \binits{Z.}},
\bauthor{\bsnm{Smart}, \binits{M.A.}},
\bauthor{\bsnm{Chaudhuri}, \binits{K.}},
\bauthor{\bsnm{Zou}, \binits{J.}}:
\bctitle{Approximate data deletion from machine learning models}.
In: \bbtitle{International Conference on Artificial Intelligence and
  Statistics},
pp. \bfpage{2008}--\blpage{2016}
(\byear{2021}).
\bcomment{PMLR}
\end{bchapter}
\endbibitem

\bibitem{neel2021descent}
\begin{bchapter}
\bauthor{\bsnm{Neel}, \binits{S.}},
\bauthor{\bsnm{Roth}, \binits{A.}},
\bauthor{\bsnm{Sharifi-Malvajerdi}, \binits{S.}}:
\bctitle{Descent-to-delete: Gradient-based methods for machine unlearning}.
In: \bbtitle{Algorithmic Learning Theory},
pp. \bfpage{931}--\blpage{962}
(\byear{2021}).
\bcomment{PMLR}
\end{bchapter}
\endbibitem

\bibitem{thudi2021unrolling}
\begin{botherref}
\oauthor{\bsnm{Thudi}, \binits{A.}},
\oauthor{\bsnm{Deza}, \binits{G.}},
\oauthor{\bsnm{Chandrasekaran}, \binits{V.}},
\oauthor{\bsnm{Papernot}, \binits{N.}}:
Unrolling sgd: Understanding factors influencing machine unlearning.
arXiv preprint arXiv:2109.13398
(2021)
\end{botherref}
\endbibitem

\bibitem{warnecke2021machine}
\begin{botherref}
\oauthor{\bsnm{Warnecke}, \binits{A.}},
\oauthor{\bsnm{Pirch}, \binits{L.}},
\oauthor{\bsnm{Wressnegger}, \binits{C.}},
\oauthor{\bsnm{Rieck}, \binits{K.}}:
Machine unlearning of features and labels.
arXiv preprint arXiv:2108.11577
(2021)
\end{botherref}
\endbibitem

\bibitem{he2021deepobliviate}
\begin{botherref}
\oauthor{\bsnm{He}, \binits{Y.}},
\oauthor{\bsnm{Meng}, \binits{G.}},
\oauthor{\bsnm{Chen}, \binits{K.}},
\oauthor{\bsnm{He}, \binits{J.}},
\oauthor{\bsnm{Hu}, \binits{X.}}:
Deepobliviate: A powerful charm for erasing data residual memory in deep neural
  networks.
arXiv preprint arXiv:2105.06209
(2021)
\end{botherref}
\endbibitem

\bibitem{gong2021forget}
\begin{botherref}
\oauthor{\bsnm{Gong}, \binits{J.}},
\oauthor{\bsnm{Simeone}, \binits{O.}},
\oauthor{\bsnm{Kassab}, \binits{R.}},
\oauthor{\bsnm{Kang}, \binits{J.}}:
Forget-svgd: Particle-based bayesian federated unlearning.
arXiv preprint arXiv:2111.12056
(2021)
\end{botherref}
\endbibitem

\bibitem{guo2022vertical}
\begin{botherref}
\oauthor{\bsnm{Guo}, \binits{T.}},
\oauthor{\bsnm{Guo}, \binits{S.}},
\oauthor{\bsnm{Zhang}, \binits{J.}},
\oauthor{\bsnm{Xu}, \binits{W.}},
\oauthor{\bsnm{Wang}, \binits{J.}}:
Vertical machine unlearning: Selectively removing sensitive information from
  latent feature space.
arXiv preprint arXiv:2202.13295
(2022)
\end{botherref}
\endbibitem

\bibitem{cauwenberghs2000incremental}
\begin{botherref}
\oauthor{\bsnm{Cauwenberghs}, \binits{G.}},
\oauthor{\bsnm{Poggio}, \binits{T.}}:
Incremental and decremental support vector machine learning.
Advances in neural information processing systems
\textbf{13}
(2000)
\end{botherref}
\endbibitem

\bibitem{tsai2014incremental}
\begin{bchapter}
\bauthor{\bsnm{Tsai}, \binits{C.-H.}},
\bauthor{\bsnm{Lin}, \binits{C.-Y.}},
\bauthor{\bsnm{Lin}, \binits{C.-J.}}:
\bctitle{Incremental and decremental training for linear classification}.
In: \bbtitle{Proceedings of the 20th ACM SIGKDD International Conference on
  Knowledge Discovery and Data Mining},
pp. \bfpage{343}--\blpage{352}
(\byear{2014})
\end{bchapter}
\endbibitem

\bibitem{karasuyama2010multiple}
\begin{barticle}
\bauthor{\bsnm{Karasuyama}, \binits{M.}},
\bauthor{\bsnm{Takeuchi}, \binits{I.}}:
\batitle{Multiple incremental decremental learning of support vector machines}.
\bjtitle{IEEE Transactions on Neural Networks}
\bvolume{21}(\bissue{7}),
\bfpage{1048}--\blpage{1059}
(\byear{2010})
\end{barticle}
\endbibitem

\bibitem{kearns1998efficient}
\begin{barticle}
\bauthor{\bsnm{Kearns}, \binits{M.}}:
\batitle{Efficient noise-tolerant learning from statistical queries}.
\bjtitle{Journal of the ACM (JACM)}
\bvolume{45}(\bissue{6}),
\bfpage{983}--\blpage{1006}
(\byear{1998})
\end{barticle}
\endbibitem

\bibitem{martens2014new}
\begin{botherref}
\oauthor{\bsnm{Martens}, \binits{J.}}:
New insights and perspectives on the natural gradient method.
arXiv preprint arXiv:1412.1193
(2014)
\end{botherref}
\endbibitem

\bibitem{dwork2014algorithmic}
\begin{barticle}
\bauthor{\bsnm{Dwork}, \binits{C.}},
\bauthor{\bsnm{Roth}, \binits{A.}}, \betal:
\batitle{The algorithmic foundations of differential privacy.}
\bjtitle{Found. Trends Theor. Comput. Sci.}
\bvolume{9}(\bissue{3-4}),
\bfpage{211}--\blpage{407}
(\byear{2014})
\end{barticle}
\endbibitem

\bibitem{chaudhuri2008privacy}
\begin{botherref}
\oauthor{\bsnm{Chaudhuri}, \binits{K.}},
\oauthor{\bsnm{Monteleoni}, \binits{C.}}:
Privacy-preserving logistic regression.
Advances in neural information processing systems
\textbf{21}
(2008)
\end{botherref}
\endbibitem

\bibitem{golatkar2020forgetting}
\begin{bchapter}
\bauthor{\bsnm{Golatkar}, \binits{A.}},
\bauthor{\bsnm{Achille}, \binits{A.}},
\bauthor{\bsnm{Soatto}, \binits{S.}}:
\bctitle{Forgetting outside the box: Scrubbing deep networks of information
  accessible from input-output observations}.
In: \bbtitle{European Conference on Computer Vision},
pp. \bfpage{383}--\blpage{398}
(\byear{2020}).
\bcomment{Springer}
\end{bchapter}
\endbibitem

\bibitem{koh2017understanding}
\begin{bchapter}
\bauthor{\bsnm{Koh}, \binits{P.W.}},
\bauthor{\bsnm{Liang}, \binits{P.}}:
\bctitle{Understanding black-box predictions via influence functions}.
In: \bbtitle{International Conference on Machine Learning},
pp. \bfpage{1885}--\blpage{1894}
(\byear{2017}).
\bcomment{PMLR}
\end{bchapter}
\endbibitem

\bibitem{giordano2019swiss}
\begin{bchapter}
\bauthor{\bsnm{Giordano}, \binits{R.}},
\bauthor{\bsnm{Stephenson}, \binits{W.}},
\bauthor{\bsnm{Liu}, \binits{R.}},
\bauthor{\bsnm{Jordan}, \binits{M.}},
\bauthor{\bsnm{Broderick}, \binits{T.}}:
\bctitle{A swiss army infinitesimal jackknife}.
In: \bbtitle{The 22nd International Conference on Artificial Intelligence and
  Statistics},
pp. \bfpage{1139}--\blpage{1147}
(\byear{2019}).
\bcomment{PMLR}
\end{bchapter}
\endbibitem

\bibitem{zhang2017diagnosing}
\begin{bchapter}
\bauthor{\bsnm{Zhang}, \binits{Z.}},
\bauthor{\bsnm{Sparks}, \binits{E.R.}},
\bauthor{\bsnm{Franklin}, \binits{M.J.}}:
\bctitle{Diagnosing machine learning pipelines with fine-grained lineage}.
In: \bbtitle{Proceedings of the 26th International Symposium on
  High-Performance Parallel and Distributed Computing},
pp. \bfpage{143}--\blpage{153}
(\byear{2017})
\end{bchapter}
\endbibitem

\bibitem{luo2021roadmap}
\begin{barticle}
\bauthor{\bsnm{Luo}, \binits{G.}}, \betal:
\batitle{A roadmap for automating lineage tracing to aid automatically
  explaining machine learning predictions for clinical decision support}.
\bjtitle{JMIR Medical Informatics}
\bvolume{9}(\bissue{5}),
\bfpage{27778}
(\byear{2021})
\end{barticle}
\endbibitem

\bibitem{thiago2020managing}
\begin{bchapter}
\bauthor{\bsnm{Thiago}, \binits{R.M.}},
\bauthor{\bsnm{Souza}, \binits{R.}},
\bauthor{\bsnm{Azevedo}, \binits{L.}},
\bauthor{\bsnm{Soares}, \binits{E.F.D.S.}},
\bauthor{\bsnm{Santos}, \binits{R.}},
\bauthor{\bsnm{Dos~Santos}, \binits{W.}},
\bauthor{\bsnm{De~Bayser}, \binits{M.}},
\bauthor{\bsnm{Cardoso}, \binits{M.C.}},
\bauthor{\bsnm{Moreno}, \binits{M.F.}},
\bauthor{\bsnm{Cerqueira}, \binits{R.}}:
\bctitle{Managing data lineage of o\&g machine learning models: the sweet spot
  for shale use case}.
In: \bbtitle{First EAGE Digitalization Conference and Exhibition},
vol. \bseriesno{2020},
pp. \bfpage{1}--\blpage{5}
(\byear{2020}).
\bcomment{European Association of Geoscientists \& Engineers}
\end{bchapter}
\endbibitem

\bibitem{li2022making}
\begin{botherref}
\oauthor{\bsnm{Li}, \binits{Y.}},
\oauthor{\bsnm{Zheng}, \binits{X.}},
\oauthor{\bsnm{Chen}, \binits{C.}},
\oauthor{\bsnm{Liu}, \binits{J.}}:
Making recommender systems forget: Learning and unlearning for erasable
  recommendation.
arXiv preprint arXiv:2203.11491
(2022)
\end{botherref}
\endbibitem

\bibitem{shokri2017membership}
\begin{bchapter}
\bauthor{\bsnm{Shokri}, \binits{R.}},
\bauthor{\bsnm{Stronati}, \binits{M.}},
\bauthor{\bsnm{Song}, \binits{C.}},
\bauthor{\bsnm{Shmatikov}, \binits{V.}}:
\bctitle{Membership inference attacks against machine learning models}.
In: \bbtitle{2017 IEEE Symposium on Security and Privacy (SP)},
pp. \bfpage{3}--\blpage{18}
(\byear{2017}).
\bcomment{IEEE}
\end{bchapter}
\endbibitem

\bibitem{yeom2018privacy}
\begin{bchapter}
\bauthor{\bsnm{Yeom}, \binits{S.}},
\bauthor{\bsnm{Giacomelli}, \binits{I.}},
\bauthor{\bsnm{Fredrikson}, \binits{M.}},
\bauthor{\bsnm{Jha}, \binits{S.}}:
\bctitle{Privacy risk in machine learning: Analyzing the connection to
  overfitting}.
In: \bbtitle{2018 IEEE 31st Computer Security Foundations Symposium (CSF)},
pp. \bfpage{268}--\blpage{282}
(\byear{2018}).
\bcomment{IEEE}
\end{bchapter}
\endbibitem

\bibitem{sablayrolles2019white}
\begin{bchapter}
\bauthor{\bsnm{Sablayrolles}, \binits{A.}},
\bauthor{\bsnm{Douze}, \binits{M.}},
\bauthor{\bsnm{Schmid}, \binits{C.}},
\bauthor{\bsnm{Ollivier}, \binits{Y.}},
\bauthor{\bsnm{J{\'e}gou}, \binits{H.}}:
\bctitle{White-box vs black-box: Bayes optimal strategies for membership
  inference}.
In: \bbtitle{International Conference on Machine Learning},
pp. \bfpage{5558}--\blpage{5567}
(\byear{2019}).
\bcomment{PMLR}
\end{bchapter}
\endbibitem

\bibitem{hayes2019logan}
\begin{bchapter}
\bauthor{\bsnm{Hayes}, \binits{J.}},
\bauthor{\bsnm{Melis}, \binits{L.}},
\bauthor{\bsnm{Danezis}, \binits{G.}},
\bauthor{\bsnm{De~Cristofaro}, \binits{E.}}:
\bctitle{Logan: Membership inference attacks against generative models}.
In: \bbtitle{Proceedings on Privacy Enhancing Technologies (PoPETs)},
vol. \bseriesno{2019},
pp. \bfpage{133}--\blpage{152}
(\byear{2019}).
\bcomment{De Gruyter}
\end{bchapter}
\endbibitem

\end{thebibliography}


\end{document}